\documentclass[lettersize,journal]{IEEEtran}
\usepackage{amsmath,amsfonts}
\usepackage{algorithmic}
\usepackage{algorithm}
\usepackage{array}
\usepackage[caption=false,font=normalsize,labelfont=sf,textfont=sf]{subfig}
\usepackage{textcomp}
\usepackage{stfloats}
\usepackage{url}
\usepackage{verbatim}
\usepackage{graphicx}
\usepackage{cite}
\usepackage{amsmath,amssymb,amsfonts}
\usepackage{subcaption}
\usepackage{multirow}
\usepackage{booktabs}
\usepackage{textcomp}
\usepackage{xcolor}
\usepackage{makecell}
\begin{document}
\title{Towards High-Precision Depth Sensing via Monocular-Aided iToF and RGB Integration }
\author{Yansong Du, Yutong Deng, Yuting Zhou, Feiyu Jiao, Jian Song,~\IEEEmembership{Fellow,~IEEE,} Xun Guan,\IEEEmembership{Member,~IEEE}
        % <-this % stops a space
\thanks{Yansong Du and Yutong Deng contributed equally to this work. (Corresponding authors: Jian Song; Xun Guan.)}}

\markboth{Journal of \LaTeX\ Class Files,~Vol.~14, No.~8, August~2021}%
{Shell \MakeLowercase{\textit{et al.}}: A Sample Article Using IEEEtran.cls for IEEE Journals}

\maketitle

\begin{abstract}
This paper presents a novel iToF-RGB fusion framework designed to address the inherent limitations of indirect Time-of-Flight (iToF) depth sensing, such as low spatial resolution, limited field-of-view (FoV), and structural distortion in complex scenes. The proposed method first reprojects the narrow-FoV iToF depth map onto the wide-FoV RGB coordinate system through a precise geometric calibration and alignment module, ensuring pixel-level correspondence between modalities. A dual-encoder fusion network is then employed to jointly extract complementary features from the reprojected iToF depth and RGB image, guided by monocular depth priors to recover fine-grained structural details and perform depth super-resolution. By integrating cross-modal structural cues and depth consistency constraints, our approach achieves enhanced depth accuracy, improved edge sharpness, and seamless FoV expansion. Extensive experiments on both synthetic and real-world datasets demonstrate that the proposed framework significantly outperforms state-of-the-art methods in terms of accuracy, structural consistency, and visual quality. 
\end{abstract}


\section{Introduction}
\begin{figure*}[htbp]
    \centering
    \includegraphics[width=0.95\linewidth]{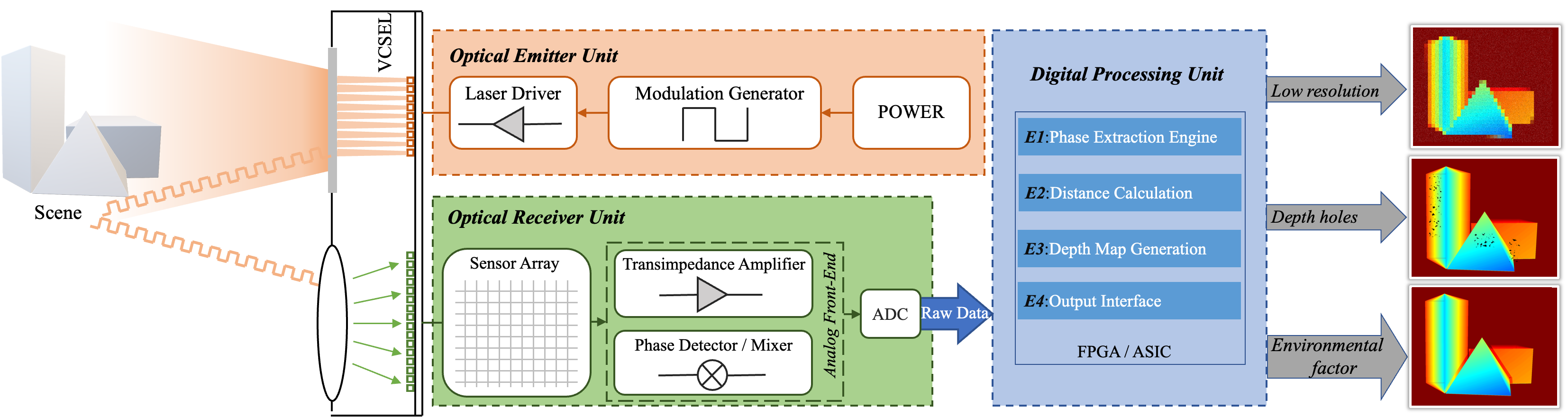}
    \caption{\footnotesize System Architecture and Challenges of an Indirect Time-of-Flight (iToF) Imaging Pipeline .}
    \label{fig:fig1}
\end{figure*}

\IEEEPARstart{I}{ndirect} Time-of-Flight (iToF) cameras acquire absolute depth information of a scene by modulating the emitted light with a continuous wave and measuring the phase shift of the reflected signal \cite{Du:25}. Owing to their compact size, low power consumption, and cost efficiency, iToF sensors have been widely deployed in smartphones, augmented reality (AR) devices, wearables, and robotic vision systems , playing an essential role in real-time perception tasks such as 3D reconstruction, gesture recognition, and autonomous navigation \cite{qiao2024rgb}. As shown in Fig.~\ref{fig:fig1}, iToF depth maps still suffer from relatively low spatial resolution, limited field of view (FoV), and a high sensitivity to noise. Non-systematic errors such as multipath interference (MPI) \cite{du2025multipath}, flying pixels, and phase wrapping can cause severe depth distortions and boundary inaccuracies, while conventional calibration and filtering methods can only partially mitigate these artifacts \cite{son2016learning,su2018deep,marco2017deeptof}. In scenes with transparent or highly reflective surfaces, depth maps are prone to holes and erroneous estimates, which limit the applicability of iToF in high-precision tasks.  
To address these limitations, monocular depth estimation (MDE)  based on RGB images has advanced rapidly with the development of deep learning \cite{masoumian2022monocular}. By learning from large-scale datasets with rich texture and geometric cues, MDE models can predict high-resolution relative depth maps and offer a wider FoV. Nevertheless, MDE lacks absolute scale information and often struggles to generalize under challenging conditions such as illumination variations and occlusions\cite{zhao2020monocular}. Combining iToF and RGB has thus emerged as a natural solution: iToF provides absolute depth as a metric reference, while RGB images contribute high-resolution structural and texture information\cite{b8}. The fusion of these two modalities can not only expand the FoV but also significantly enhance depth accuracy and robustness to noise\cite{ma2025ligo}.  In this context, Guided Depth Super-Resolution (GDSR) has become a prominent research direction for improving iToF depth quality. By fusing cross-modal features, GDSR leverages high-resolution RGB textures to refine low-resolution iToF depth maps and substantially improves both edge sharpness and structural consistency. Empirical studies have demonstrated the effectiveness of GDSR in applications such as AR/VR rendering, robotic navigation, and 3D scene understanding. As GDSR has become the mainstream approach for iToF-RGB fusion, research efforts have increasingly focused on how to achieve better cross-modal alignment and more precise depth reconstruction\cite{li2024ps5}. 

The early development of GDSR primarily relied on convolutional neural networks (CNNs), which enhanced depth quality through multi-scale feature extraction and fusion. Yan et al.  introduced a multi-scale convolutional network incorporating joint correlation learning (JCL) and iterative cross modules (IC) to achieve joint depth completion and super-resolution\cite{b9}. Building on this idea, Song et al.  integrated residual learning and channel attention modules to further improve the recovery of fine details along edges and textures\cite{b10}. Yang et al.  designed the MRG-PS dual-stream architecture, which utilized recursive guidance and staged residual learning to enhance both overall reconstruction accuracy and stability\cite{b11}. As task complexity and real-time requirements increased, the field gradually transitioned toward more powerful architectures. DELTAR (ECCV 2022) proposed an efficient high-resolution reconstruction scheme tailored for lightweight ToF sensors, while Sun et al. incorporated cross-frame consistency constraints to improve depth recovery in video sequences\cite{li2022deltar}. Metzger et al. adopted anisotropic diffusion strategies to better preserve structural fidelity under complex scenes\cite{metzger2023guided}. More recently, Zhang et al. introduced the CDF-DSR framework, which leverages continuous depth field modeling and cross-modal geometric consistency loss to achieve high-accuracy reconstruction under unsupervised settings, marking a new step toward improved precision and generalization in GDSR\cite{zhang2025cdf}.

Despite these advancements, most existing GDSR methods still rely on simple concatenation of RGB and depth features without explicitly modeling the structural discrepancies and noise characteristics between the two modalities. This often results in texture-copy artifacts, boundary inconsistencies, and insufficient robustness to complex noise. To overcome these challenges, we propose an end-to-end iToF-RGB fusion framework that performs accurate cross-modal alignment via joint calibration and geometric reprojection. A dual-branch RGB-D fusion network is designed to generate high-resolution depth predictions with strong noise suppression capability. In addition, a depth structure distillation module is incorporated, leveraging relative depth priors from MDE to enforce structural consistency and edge fidelity. Extensive experiments on both synthetic and real-world datasets validate the effectiveness of the proposed approach, showing superior performance in terms of depth accuracy, noise robustness, and structural preservation, thereby enabling high-quality depth reconstruction in challenging environments.

\section{Principle}
\subsection{Fundamentals of iToF Depth Measurement }
Indirect Time-of-Flight (iToF) cameras estimate object distance by measuring the phase delay of a modulated light signal. A continuous-wave modulated light is emitted from the transmitter, reflected by the object surface, and then received by the sensor. The phase shift  $\phi$  between the emitted and received signals is proportional to the optical path length, and the ideal distance between the object and the camera can be expressed as\cite{Du:25}:
\begin{equation}
d = \frac{c}{4 \pi f_m} \cdot \phi \tag{1}
\end{equation}
where  $c$  is the speed of light and  $f_{m}$ is the modulation frequency. In practice, however, the measured depth  $d_{m}$ deviates from the true distance  $d_{t}$ due to various factors, such as ambient light interference, sensor noise, modulation and sampling misalignment, electronic readout noise, and quantization errors. The resulting error is defined as:
\begin{equation}
\Delta d = d_m - d_t \tag{2}
\end{equation}
These error sources collectively affect the reliability of depth estimation and lead to unstable measurements and the loss of structural details. Furthermore, due to hardware limitations, iToF cameras typically provide low-resolution depth maps, which are prone to jagged edges, blurring, and local holes, making them insufficient for high-precision scene perception. Therefore, compensating for  $\Delta d$  and enhancing spatial resolution are critical challenges in iToF depth reconstruction. To address these issues, this study proposes an RGB-assisted cross-modal fusion approach that leverages the high-resolution texture and structural priors of RGB images to effectively mitigate measurement errors and improve the quality of iToF depth maps. 

\subsection{Calibration and Depth Reprojection}
To achieve accurate spatial alignment between the iToF depth map and the RGB image, we first perform a joint calibration of the iToF–RGB dual-camera system. During this process, the intrinsic matrices  $ \mathbf{\textit{K}_{\mathrm{iToF}}} $  and  $ \mathbf{\textit{K}_{\mathrm{RGB}}} $ are estimated to describe the pinhole camera parameters such as focal lengths, principal point positions, and pixel scaling factors. In addition, the extrinsic parameters  $ \mathbf{\textit{R}_{\mathrm{iToF} \rightarrow \mathrm{RGB}}} $  and $ \mathbf{\textit{t}_{\mathrm{iToF} \rightarrow \mathrm{RGB}}} $  are determined to define the rigid transformation between the iToF and RGB camera coordinate systems. Since the RGB camera often exhibits lens distortions, including radial and tangential distortions, we correct the RGB image using the distortion coefficients obtained from calibration, ensuring geometric consistency before performing reprojection. 

\begin{figure}[htbp]
  \centering
  \includegraphics[width=0.48\textwidth]{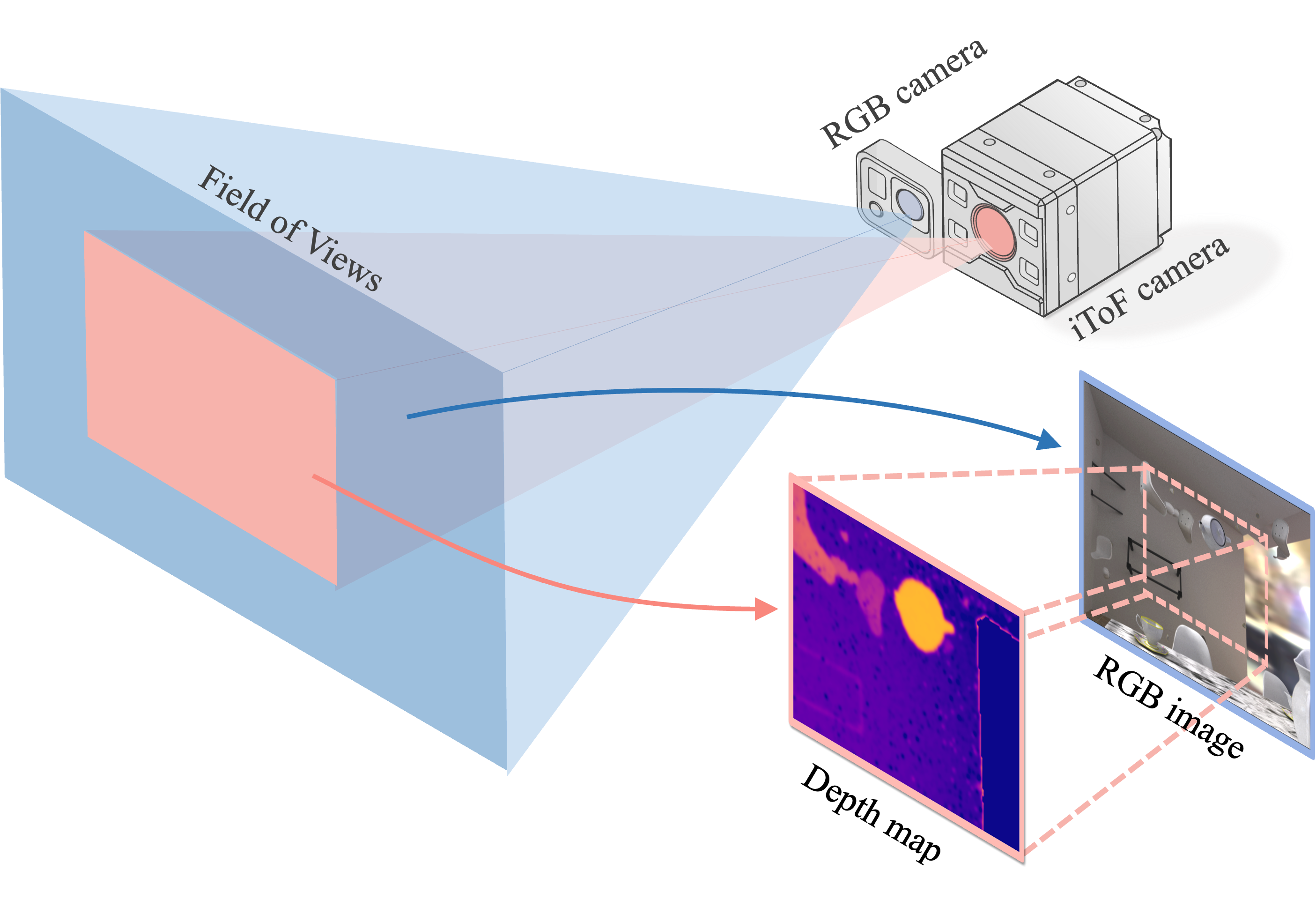}
  \caption{\footnotesize Schematic diagram of the RGB-iToF dual-camera system structure.}
  \label{fig:fig2}
\end{figure}

In the geometric reprojection stage, the pixels of the iToF depth map are mapped onto the RGB image plane to achieve pixel-level alignment. This process can be divided into three sequential steps. (1) Back-projection: each 2D pixel $\mathbf{\textit{p}_{\mathrm{iToF}}} = [\mathrm{u, v, 1}]^\mathrm{T}$ in the iToF depth map, combined with its depth $ \mathbf{\textit{Z}_{\mathrm{iToF}}} $,  is back-projected into a 3D point in the iToF camera coordinate system via $\mathbf{\textit{K}}_{\mathrm{iToF}}^{-1}$. (2) Coordinate transformation: the 3D point is transformed into the RGB camera coordinate system using the extrinsic parameters $ \mathbf{\textit{R}_{\mathrm{iToF} \rightarrow \mathrm{RGB}}} $ and $ \mathbf{\textit{t}_{\mathrm{iToF} \rightarrow \mathrm{RGB}}} $. (3) Projection and distortion mapping: the transformed point is projected onto the RGB image plane using the intrinsic matrix $ \mathbf{\textit{K}_{\mathrm{RGB}}} $, and the final pixel coordinate is obtained by applying the distortion mapping function $D(\cdot)$. The overall reprojection relationship is given by: 
\begin{equation}
\begin{aligned}
\mathbf{p}_{\mathrm{RGB}} = D \Big(
& \mathbf{K}_{\mathrm{RGB}} \Big(
\mathbf{R}_{\mathrm{iToF} \rightarrow \mathrm{RGB}}
\big(
Z_{\mathrm{iToF}} \mathbf{K}_{\mathrm{iToF}}^{-1} \mathbf{p}_{\mathrm{iToF}}
\big) \\
& + \mathbf{t}_{\mathrm{iToF} \rightarrow \mathrm{RGB}}
\Big) \Big).
\end{aligned}
\tag{3}
\end{equation}
By following these three steps and incorporating distortion correction as shown in Fig.\ref{fig:fig2}, the iToF depth map is accurately reprojected into the RGB view, enabling precise cross-modal alignment and providing a reliable foundation for subsequent feature fusion.

\subsection{MDE-Guided Enhancement of iToF Depth Maps}
The iToF depth maps often suffer from low resolution, high noise levels, and local missing regions, making them inadequate for high-precision scene perception. To address these limitations, we incorporate monocular depth estimation (MDE) as a structural prior to enhance the completeness and detail representation of depth maps. The MDE model predicts relative depth distributions based on the high-resolution texture information of RGB images, thereby compensating for the deficiencies of iToF in edge sharpness and fine geometric structures. Benefiting from its training on large-scale and diverse datasets, MDE provides stable and reliable structural cues. For cross-modal fusion, the relative depth map produced by MDE is first aligned in scale with the reprojected iToF absolute depth map to ensure consistency in depth distribution. On this basis, the MDE depth map not only fills the holes and low-confidence regions in the iToF depth map but also guides edge restoration and geometric detail refinement during feature fusion. By introducing structure-consistency constraints derived from MDE in the training process, the network effectively preserves the absolute depth accuracy of iToF while enhancing high-frequency details, resulting in final depth outputs that are significantly improved in completeness, accuracy, and structural consistency. 

\subsection{Dataset Preparation }
While iToF cameras provide raw depth measurements, acquiring high-quality ground-truth depth requires high-precision scanning equipment and careful alignment, making large-scale data collection challenging. To overcome this limitation, we primarily use the publicly available synthetic dataset \textbf{ToF-FlyingThings3D} \cite{b13} as our training source. This dataset offers RGB images and raw depth maps rendered from the same virtual camera viewpoint. By applying geometric transformations based on calibration parameters derived from our custom iToF-RGB dual-camera setup, we generate training pairs that closely approximate the real-world imaging configuration.

In addition, a small set of \textbf{real-world RGB-iToF samples} is collected using our dual-camera system, which is integrated into the dataset to enhance domain diversity and improve model generalization. These real samples are mainly used for fine-tuning and evaluation, ensuring that the trained model performs reliably under real-world conditions.

\subsection{Network Architecture}
\begin{figure*}[htbp]
    \centering
    \includegraphics[width=0.95\linewidth]{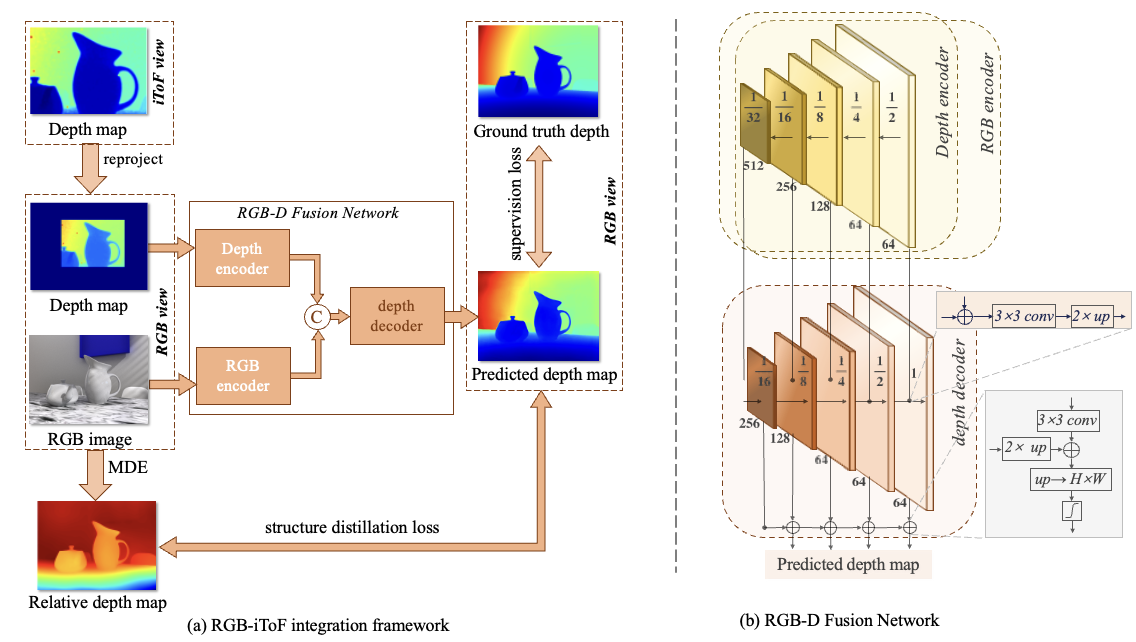}
    \caption{\footnotesize The overall framework and the architecture of the RGB-D fusion network. (a) The overall framework of the RGB-iToF integration, where the raw iToF depth is reprojected and aligned with the RGB view before being fed into the fusion network.(b) The architecture of the RGB-D fusion network, which consists of two structurally identical encoders (an RGB encoder and a depth encoder). Each encoder extracts multi-scale features, which are concatenated at each level and passed to a shared decoder. The decoder progressively fuses and upsamples the features to generate a list of depth predictions at multiple scale.}
    \label{fig:fig3}
\end{figure*}

To address the limitations of iToF cameras in practical applications—namely, low spatial resolution, narrow field-of-view (FoV), and structural distortion in complex scenes—we propose a cross-modal iToF-RGB depth reconstruction framework that integrates RGB images and monocular depth priors. Although iToF systems inherently provide reliable range measurements, their depth outputs often suffer from blurred edges, missing details, and limited spatial coverage due to sensor size and optical constraints. In contrast, RGB images offer high spatial resolution and a wider FoV, capturing rich texture and global structure. Additionally, monocular depth estimation can supply structurally consistent relative depth priors. Based on these complementary characteristics, we design an end-to-end fusion network in Fig.\ref{fig:fig3} that improves depth resolution, enhances structural details, and expands the spatial coverage, all without altering the hardware setup of the iToF system.

As illustrated in Fig. 3(a), the proposed RGB-iToF integration pipeline begins with geometric calibration and reprojection of the raw iToF depth map into the RGB camera viewpoint, ensuring pixel-level alignment across modalities. The aligned RGB image and reprojected depth map are then fed into a dual-branch fusion network, where the RGB and depth encoders independently extract modality-specific features. To explicitly enhance structural consistency and edge sharpness in the predicted depth map, a monocular depth estimation (MDE) module is incorporated during training. The MDE branch produces a relative depth prior from the RGB image, which, though lacking absolute scale, provides robust structural guidance. This prior is injected into the training process via a structure distillation loss, encouraging the network to recover clearer object boundaries and more geometrically faithful reconstructions. The entire system is trained under supervision from ground-truth depth maps in the RGB view, using both pixel-wise regression and distillation losses to jointly optimize depth accuracy and structural consistency.

Fig. 3(b) further details the architecture of the RGB-D Fusion Network. A symmetric dual-encoder structure—sharing the ResNet backbone—is adopted to extract multi-level semantic features from both RGB and depth modalities. The decoder employs a multi-scale feature fusion strategy, combining concatenation, skip connections, and pyramid-style upsampling to progressively restore the spatial structure. Each decoding stage consists of a 3×3 convolution and a 2× upsampling operation, enabling the fusion of deep contextual information with local spatial cues. This deep-to-shallow fusion mechanism ensures that both global semantics and fine-grained details are preserved throughout the reconstruction process. The final output is a high-resolution depth map that aligns with the RGB image in both resolution and FoV, offering superior accuracy, sharper edges, and improved 3D structure representation for downstream perception tasks.

\subsection{Loss Functions}
We design four complementary loss terms to jointly constrain the predicted depth map in terms of numerical accuracy, structural consistency, edge preservation, and geometric fidelity. These losses supervise the network output from the perspectives of global regression, local smoothness, structure-aware regularization, and surface normal consistency.

\textbf{Supervised Regression Loss.} Smooth L1 Loss is adopted as the main supervision signal to enhance regression stability and reduce the impact of outliers during training. This loss behaves as an L2 loss for small residuals and transitions to an L1 loss for larger errors, offering a good balance between convergence and robustness:
\begin{equation}
L_{reg} = \frac{1}{N} \sum_{i=1}^{N} \text{SmoothL1} \left( D_i^{pred} - D_i^{gt} \right)
\end{equation}
\begin{equation}
\text{SmoothL1}(x) = 
\begin{cases}
0.5\, x^{2}, & |x| < 1 \\[6pt]
|x| - 0.5, & otherwise
\end{cases}
\end{equation}
where $ D^{pred} $ denotes the predicted depth map, $ D^{gt} $ denotes the ground-truth depth map, and $ N $ is the number of valid pixels.

\textbf{Edge-aware Smoothness Loss.} Global smoothness constraints may over-blur object boundaries due to sharp depth discontinuities. An edge-aware second-order smoothness loss is introduced to mitigate this issue, where image gradients guide the smoothness weights. This formulation promotes spatial coherence in textureless regions while preserving edges: 
\begin{equation}
L_{smooth} = 
\frac{1}{2} \Big(
\mathbb{E} \big[\, w_x \, |\, \nabla_{xx} D | \big]
+
\mathbb{E} \big[\, w_y \, |\, \nabla_{yy} D | \big]
\Big)
\label{eq:smooth}
\end{equation}
\begin{equation}
w_x = \exp\!\big(- \alpha\, | \nabla_x I | \big),
w_y = \exp\!\big(- \alpha\, | \nabla_y I | \big)
\label{eq:weights}
\end{equation}
Where $I$ denotes the input RGB image, $D$ denotes the predicted depth map, and the edge-aware weights $w_x$ and $w_y$ are derived from the image gradients to adaptively smooth the depth while preserving image edges.

\textbf{Structure Distillation Loss.} The predicted depth map is regularized to match the structure of the monocular relative depth map through a structure distillation mechanism. The Structural Similarity Index Measure (SSIM) is employed to quantify local structural agreement, considering luminance, contrast, and correlation within image patches: 
\begin{equation}
L_{struct} = \frac{1}{2} 
\Big(
1 - \text{SSIM}(x, y)
\Big)
\label{eq:struct_loss}
\end{equation}
\begin{equation}
\text{SSIM}(x, y) = 
\frac{
(2 \mu_x \mu_y + C_1)(2 \sigma_{xy} + C_2)
}{
(\mu_x^2 + \mu_y^2 + C_1)(\sigma_x + \sigma_y + C_2)
}
\label{eq:ssim_index}
\end{equation}
where $x$ denotes the predicted depth map and $y$ denotes the monocular relative depth map,and $ \mu $, $ \sigma$ , and $ \sigma_{xy} $ represent the local mean, variance, and covariance,respectively, which are computed using a sliding window average.

\textbf{Normal Consistency Loss.} Geometric fidelity is further enforced through a normal consistency loss, which encourages the predicted depth map to align with the surface normals computed from the ground truth. This loss improves the preservation of local curvature and surface orientation: 
\begin{equation}
    L_{\text{normal}} = \frac{1}{N} \sum_{i=1}^{N} \left( 1 - \left\langle \mathbf{n}_i^{\text{pred}}, \mathbf{n}_i^{\text{gt}} \right\rangle \right)
\end{equation}
Here, $\mathbf{n}_i^{\text{pred}}$ and $\mathbf{n}_i^{\text{gt}}$ are the unit surface normals at pixel $i$, estimated from the predicted and ground truth depth maps, respectively. The dot product $\langle \cdot, \cdot \rangle$ evaluates angular alignment. Surface normals are calculated via finite differences or Sobel filters and normalized before comparison.

\textbf{Total Loss Function.} The overall training objective is formulated as a weighted combination of the four loss terms:
\begin{equation}
    L_{\text{total}} = \lambda_1 L_{\text{reg}} + \lambda_2 L_{\text{struct}} + \lambda_3 L_{\text{smooth}} + \lambda_4 L_{\text{normal}}
\end{equation}
The weighting coefficients $\lambda_1$ through $\lambda_4$ are empirically tuned to balance accuracy, structure preservation, smoothness, and geometric consistency.

\section{Experimental results and discussion}
The network proposed in this study is trained on the synthetic ToF-FlyingThings3D dataset. Using the calibration parameters of the iToF-RGB dual-camera system, we reproject the original depth map and RGB image into the respective real camera viewpoints, constructing pseudo-stereo data for training. The network adopts ResNet-18 as the encoder, and a pyramid decoder that progressively upsamples and fuses multi-scale features, ultimately generating a normalized depth map via a tanh activation function. The dataset is split into training and validation sets with a ratio of approximately 95:5. The model is trained for 100 epochs with a batch size of 4, using the Radam optimizer and an initial learning rate of \(1 \times 10^{-4}\) without learning rate decay. The loss functions include regression loss, smoothness loss, and structure distillation loss, with weights of 1.0, 0.1, and 0.01, respectively. The monocular structure distillation component leverages relative depth maps generated by the MiDaS model to guide structural consistency~\cite{b12}.
\begin{table}[htbp]
  \centering
  \caption{\textsc{Quantitative Evaluation Metrics of Depth Prediction Across Different Regions}}
  \label{tab:eval_metrics}
  \renewcommand\arraystretch{1.68}
  \begin{tabular}{|>{\centering\arraybackslash}m{1.8cm}|c|c|c|}
    \hline
    \multirow{2}{*}{\centering\textbf{Metric}} & \multicolumn{3}{c|}{\textbf{Evaluation Regions}} \\
    \cline{2-4}
    & \textbf{\textit{Full Image}} 
    & \makecell{\rule{0pt}{8.5pt} \textbf{\textit{Outside}} \\ \textbf{\textit{iToF FOV}}}
    & \makecell{\rule{0pt}{8.5pt} \textbf{\textit{Overlapping}} \\ \textbf{\textit{iToF FOV}}}  \\
    \hline
    MAE $\downarrow$     & 1.2905 & 1.3043 & 1.2544 \\
    \hline
    MSE $\downarrow$     & 9.6092 & 9.5439 & 9.7323 \\
    \hline
    RMSE $\downarrow$    & 2.9862 & 2.9397 & 3.0389 \\
    \hline
    AbsRel $\downarrow$  & 0.0190 & 0.0196 & 0.0175 \\
    \hline
    $\delta_1\uparrow$   & 0.9923 & 0.9920 & 0.9930 \\
    \hline
    $\delta_2\uparrow$   & 0.9978 & 0.9977 & 0.9980 \\
    \hline
    $\delta_3\uparrow$   & 0.9991 & 0.9990 & 0.9994 \\
    \hline
  \end{tabular}
\end{table}

We evaluated the trained model on the synthetic dataset using several standard quantitative metrics. In particular, $\delta_1, \delta_2, \text{ and } \delta_3$ represent the proportion of pixels where the ratio between the predicted depth and ground truth depth satisfies $\max(D/\hat{D}, \hat{D}/D)  < 1.25,1.25^2,1.25^3$ , respectively, reflecting the model’s accuracy coverage under different tolerance thresholds. The experimental results show that all metrics perform well, especially $\delta_1$ , which exceeds 0.99, indicating high prediction accuracy for the vast majority of pixels. Furthermore, we partitioned the testing area into regions Outside and Overlapping the iToF FOV. Experimental results show that the MAE, AbsRel, and $\delta$ metrics in the Overlapping iToF FOV region outperform those in the Outside iToF FOV region. This indicates that in the overlapping field of view, the fused depth map can better leverage the complementary advantages of iToF depth and RGB information, thereby improving prediction accuracy and robustness. In contrast, the performance in the Outside iToF FOV region slightly decreases due to the lack of iToF depth input. The overall evaluation results are detailed in Table~\ref{tab:eval_metrics}.
\begin{figure}[htbp]
  \centering
  \includegraphics[width=0.48\textwidth]{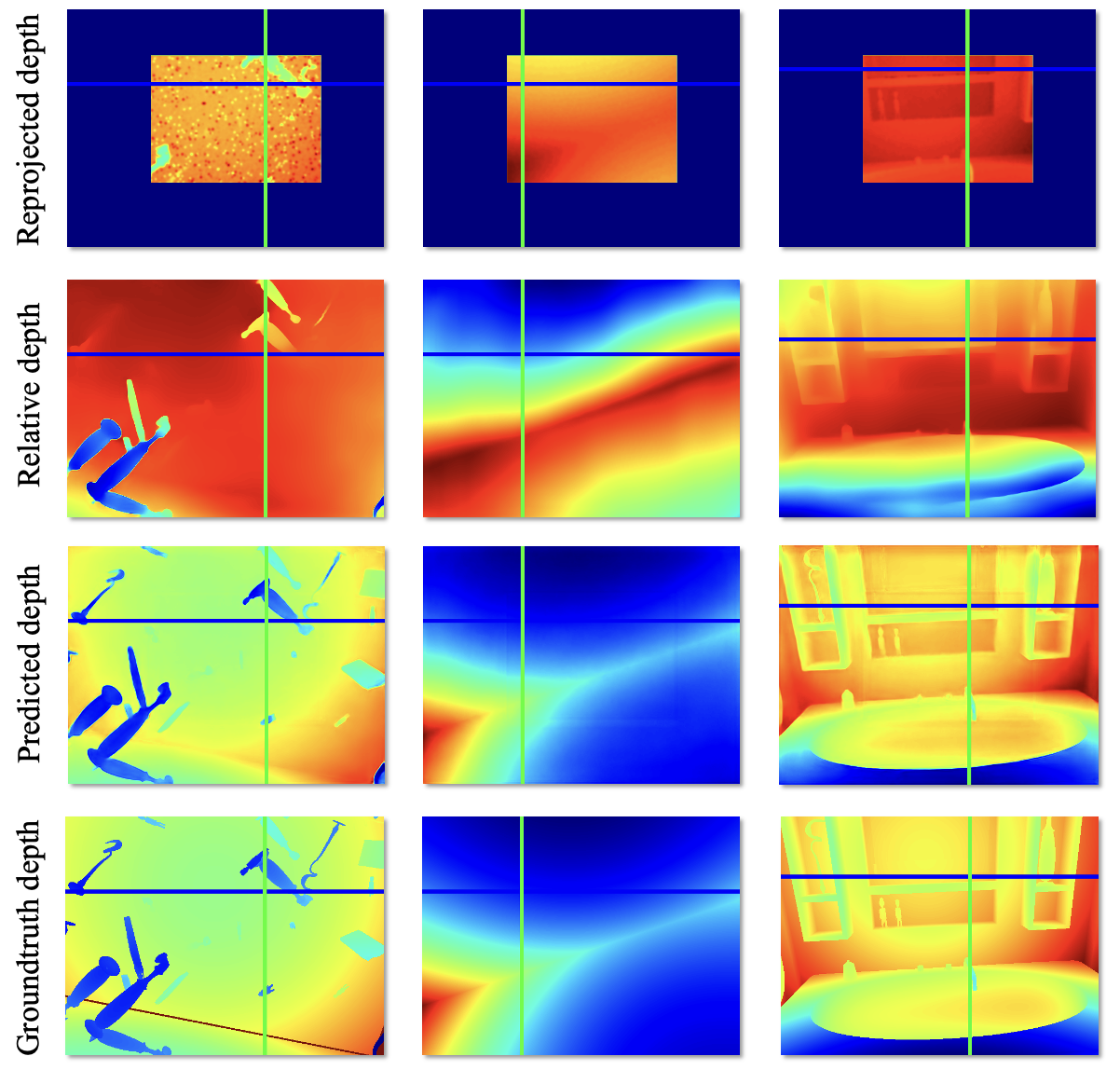}
  \caption{\footnotesize Depth map comparison including reprojected depth, monocular relative depth, predicted depth, and ground truth depth.}
  \label{fig:fig4}
\end{figure}
Subsequently, we utilized our self-built iToF-RGB dual-camera system to capture image samples from three representative real-world scenarios, aiming to visualize the depth estimation error distribution across different regions and to evaluate the generalization ability of our model in complex environments. The corresponding depth error curves are presented in Fig.\ref{fig:fig4}. The first scenario demonstrates obvious holes and noise in the raw iToF depth maps, caused by multipath interference or low surface reflectivity. The second scenario highlights deviations in monocular depth estimation within texture-sparse regions. The third scenario corresponds to areas with depth discontinuities and object boundaries. In these challenging regions, the iToF depth maps tend to suffer from missing data or large deviations, while monocular methods often produce incorrect predictions due to insufficient visual cues. By integrating structural information from RGB images and geometric priors from iToF data, the proposed fusion network effectively completes missing regions, accurately restores boundary details, and significantly suppresses errors, resulting in depth maps with improved structural consistency and more accurate scale alignment.  

\begin{figure}[htbp]
  \centering
  \includegraphics[width=0.48\textwidth]{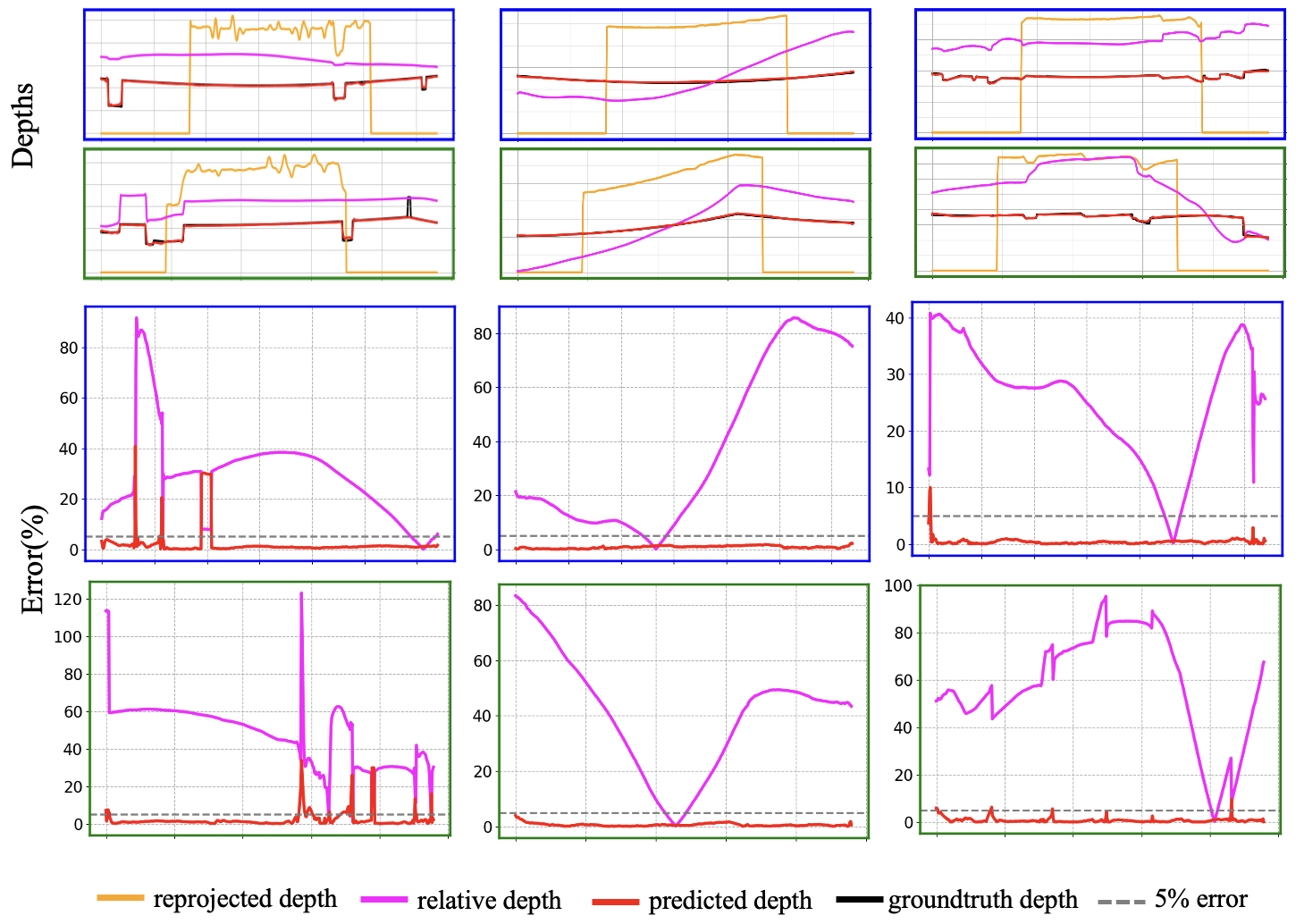}
  \caption{\footnotesize Cross-sectional depth profiles and corresponding error curves in the horizontal and vertical directions, combined with comparisons among the reprojected depth map, monocular estimation, predicted result, and ground truth, visually demonstrate local depth differences and structural deviations. }
  \label{fig:fig5}
\end{figure}

The Fig.\ref{fig:fig5} presents the quantitative results, further demonstrating the robustness and generalization capability of the proposed model in complex real-world scenarios. In each image, we randomly selected one row and one column of pixels to compare the predicted depth values with the ground truth along the selected lines, enabling a more detailed analysis of the error distribution. Experimental results show that the error rates of our predictions are generally below 5\% compared to the ground truth, significantly outperforming the results of monocular relative estimation under raw data conditions. This highlights the remarkable advantage of our method in practical applications. 

\section{Conclusion}
In this work, we propose a cross-modal depth reconstruction method that fuses indirect Time-of-Flight (iToF) depth maps with RGB images. To address the limitations of iToF cameras, including low resolution, narrow field of view, and vulnerability to multipath interference, we design a geometric reprojection-based alignment module and a dual-branch fusion network. A structural distillation module is also incorporated to introduce structural priors from monocular depth estimation. The proposed method demonstrates strong reconstruction capability on both synthetic and real-world datasets, effectively recovering depth structures in challenging regions with sparse textures, broken edges, and areas beyond the iToF sensing range. The predicted depth maps exhibit high geometric consistency and accurate scale alignment, showing robustness and suitability for high-quality depth sensing tasks in complex environments.

\section*{Acknowledgment}
This work is financially supported by National Natural Science Foundation of China (U23A20282); Shenzhen Science, Technology and Innovation Commission (KJZD20240903095600001, JCYJ20220530142809022).

\bibliographystyle{IEEEtran}  % 选用 IEEEtran 样式
\bibliography{refs}           % refs.bib，省略后缀

\begin{thebibliography}{10}
\providecommand{\url}[1]{#1}
\csname url@samestyle\endcsname
\providecommand{\newblock}{\relax}
\providecommand{\bibinfo}[2]{#2}
\providecommand{\BIBentrySTDinterwordspacing}{\spaceskip=0pt\relax}
\providecommand{\BIBentryALTinterwordstretchfactor}{4}
\providecommand{\BIBentryALTinterwordspacing}{\spaceskip=\fontdimen2\font plus
\BIBentryALTinterwordstretchfactor\fontdimen3\font minus \fontdimen4\font\relax}
\providecommand{\BIBforeignlanguage}[2]{{%
\expandafter\ifx\csname l@#1\endcsname\relax
\typeout{** WARNING: IEEEtran.bst: No hyphenation pattern has been}%
\typeout{** loaded for the language `#1'. Using the pattern for}%
\typeout{** the default language instead.}%
\else
\language=\csname l@#1\endcsname
\fi
#2}}
\providecommand{\BIBdecl}{\relax}
\BIBdecl

\bibitem{Du:25}
Y.~Du, Z.~Jiang, J.~Tian, and X.~Guan, ``Modeling, analysis, and optimization of random error in indirect time-of-flight camera,'' \emph{Opt. Express}, vol.~33, no.~2, pp. 1983--1994, 2025.

\bibitem{qiao2024rgb}
X.~Qiao, M.~Poggi, P.~Deng, H.~Wei, C.~Ge, and S.~Mattoccia, ``Rgb guided tof imaging system: A survey of deep learning-based methods,'' \emph{International Journal of Computer Vision}, vol. 132, no.~11, pp. 4954--4991, 2024.

\bibitem{du2025multipath}
Y.~Du, Y.~Deng, Y.~Zhou \emph{et~al.}, ``Multipath interference suppression in indirect time-of-flight imaging via a novel compressed sensing framework,'' \emph{arXiv preprint arXiv:2507.19546}, 2025.

\bibitem{son2016learning}
K.~Son, M.-Y. Liu, and Y.~Taguchi, ``Learning to remove multipath distortions in time-of-flight range images for a robotic arm setup,'' in \emph{2016 IEEE International Conference on Robotics and Automation (ICRA)}.\hskip 1em plus 0.5em minus 0.4em\relax IEEE, 2016, pp. 3390--3397.

\bibitem{su2018deep}
S.~Su, F.~Heide, G.~Wetzstein, and W.~Heidrich, ``Deep end-to-end time-of-flight imaging,'' in \emph{Proceedings of the IEEE Conference on Computer Vision and Pattern Recognition}, 2018, pp. 6383--6392.

\bibitem{marco2017deeptof}
J.~Marco, Q.~Hernandez, A.~Munoz, Y.~Dong, A.~Jarabo, M.~H. Kim, X.~Tong, and D.~Gutierrez, ``Deeptof: off-the-shelf real-time correction of multipath interference in time-of-flight imaging,'' \emph{ACM Transactions on Graphics (ToG)}, vol.~36, no.~6, pp. 1--12, 2017.

\bibitem{masoumian2022monocular}
A.~Masoumian, H.~A. Rashwan, J.~Cristiano, M.~S. Asif, and D.~Puig, ``Monocular depth estimation using deep learning: A review,'' \emph{Sensors (Basel)}, vol.~22, no.~14, p. 5353, Jul 18 2022.

\bibitem{zhao2020monocular}
C.~Zhao, Q.~Sun, C.~Zhang, Y.~Tang, and F.~Qian, ``Monocular depth estimation based on deep learning: An overview,'' \emph{Science China Technological Sciences}, vol.~63, no.~9, pp. 1612--1627, 2020.

\bibitem{b8}
Z.~Zhong, X.~Liu, J.~Jiang, D.~Zhao, and X.~Ji, ``Deep attentional guided image filtering,'' \emph{IEEE Transactions on Neural Networks and Learning Systems}, 2023.

\bibitem{ma2025ligo}
Y.~Ma, N.~G. Lukhanin, E.~Wang, K.~Shum, Y.~Du, L.~Lin, and X.~Guan, ``Ligo: Llm-enhanced iterative graphic optimization for large field-of-view underwater 3d reconstruction,'' in \emph{AI and Optical Data Sciences VI}, vol. 13375.\hskip 1em plus 0.5em minus 0.4em\relax SPIE, 2025, pp. 13--23.

\bibitem{li2024ps5}
F.~Li, Y.~Liu, J.~Qi, Y.~Du, Q.~Wang, W.~Ma, X.~Xu, and Z.~Zhang, ``Ps5-net: a medical image segmentation network with multiscale resolution,'' \emph{Journal of Medical Imaging}, vol.~11, no.~1, pp. 014\,008--014\,008, 2024.

\bibitem{b9}
Z.~Yan, K.~Wang, X.~Li, Z.~Zhang, G.~Li, J.~Li, and J.~Yang, ``Learning complementary correlations for depth super-resolution with incomplete data in real world,'' \emph{IEEE transactions on neural networks and learning systems}, vol.~35, no.~4, pp. 5616--5626, 2022.

\bibitem{b10}
X.~Song, Y.~Dai, D.~Zhou, L.~Liu, W.~Li, H.~Li, and R.~Yang, ``Channel attention based iterative residual learning for depth map super-resolution,'' in \emph{Proceedings of the ieee/cvf conference on computer vision and pattern recognition}, 2020, pp. 5631--5640.

\bibitem{b11}
B.~Yang, X.~Fan, Z.~Zheng, X.~Liu, K.~Zhang, and J.~Lei, ``Depth map super-resolution via multilevel recursive guidance and progressive supervision,'' \emph{IEEE Access}, vol.~7, pp. 57\,616--57\,622, 2019.

\bibitem{li2022deltar}
Y.~Li, X.~Liu, W.~Dong, H.~Zhou, H.~Bao, G.~Zhang, Y.~Zhang, and Z.~Cui, ``Deltar: Depth estimation from a light-weight tof sensor and rgb image,'' in \emph{European conference on computer vision}.\hskip 1em plus 0.5em minus 0.4em\relax Springer, 2022, pp. 619--636.

\bibitem{metzger2023guided}
N.~Metzger, R.~C. Daudt, and K.~Schindler, ``Guided depth super-resolution by deep anisotropic diffusion,'' in \emph{Proceedings of the IEEE/CVF Conference on Computer Vision and Pattern Recognition}, 2023, pp. 18\,237--18\,246.

\bibitem{zhang2025cdf}
S.~Zhang, J.~Dong, Y.~Ma, H.~Cai, M.~Wang, Y.~Li, T.~B. Kabika, X.~Li, and W.~Hou, ``Cdf-dsr: Learning continuous depth field for self-supervised rgb-guided depth map super resolution,'' \emph{Information Fusion}, vol. 117, p. 102884, 2025.

\bibitem{b13}
D.~Qiu, J.~Pang, W.~Sun, and C.~Yang, ``Deep end-to-end alignment and refinement for time-of-flight rgb-d module,'' in \emph{Proceedings of the ieee/cvf international conference on computer vision}, 2019, pp. 9994--10\,003.

\bibitem{b12}
R.~Ranftl, K.~Lasinger, D.~Hafner, K.~Schindler, and V.~Koltun, ``Towards robust monocular depth estimation: Mixing datasets for zero-shot cross-dataset transfer,'' \emph{IEEE transactions on pattern analysis and machine intelligence}, vol.~44, no.~3, pp. 1623--1637, 2020.

\end{thebibliography}
 
\end{document}